\documentclass[10pt,twocolumn,letterpaper]{article}

\usepackage{iccv}              

\definecolor{iccvblue}{rgb}{0.21,0.49,0.74}
\usepackage[pagebackref,breaklinks,colorlinks,allcolors=iccvblue]{hyperref}

\def\paperID{6975} 
\def\confName{ICCV}
\def\confYear{2025}
\usepackage[accsupp]{axessibility}
\usepackage{multirow}

\def\cc{\cellcolor[RGB]{230,230,230}} 
\usepackage{pifont}
\newcommand{\cmark}{\ding{51}}%
\newcommand{\xmark}{\ding{55}}%

\title{Describe, Adapt and Combine: Empowering CLIP Encoders for Open-set 3D Object Retrieval}

\author{
Zhichuan Wang\textsuperscript{1}\quad
Yang Zhou\textsuperscript{2}\quad
Zhe Liu\textsuperscript{3}\quad
Rui Yu\textsuperscript{4}\quad
Song Bai\textsuperscript{5}\\
Yulong Wang\textsuperscript{1}\thanks{Corresponding author}\quad
Xinwei He\textsuperscript{1}{\raisebox{0.43ex}{\footnotesize *}}
\quad
Xiang Bai\textsuperscript{6}\\
\textsuperscript{1}Huazhong Agricultural University\quad \textsuperscript{2}Shenzhen University \quad
\textsuperscript{3}The University of Hong Kong\\
\textsuperscript{4}University of Louisville\quad 
\textsuperscript{5}ByteDance\quad 
\textsuperscript{6}Huazhong University of Science and Technology\\
{\tt\small wzc\_65@webmail.hzau.edu.cn, xwhe@hzau.edu.cn}
}

\begin{document}
\maketitle

\begin{abstract}

Open-set 3D object retrieval (3DOR) is an emerging task aiming to retrieve 3D objects of unseen categories beyond the training set. Existing methods typically utilize all modalities (\emph{i.e.}, voxels, point clouds, multi-view images) and train specific backbones before fusion. However, they still struggle to produce generalized representations due to insufficient 3D training data. Being contrastively pre-trained on web-scale image-text pairs, CLIP inherently produces generalized representations for a wide range of downstream tasks. Building upon it, we present a simple yet effective framework named \textbf{D}escribe, \textbf{A}dapt and \textbf{C}ombine (\textbf{DAC}) by taking only multi-view images for open-set 3DOR.
DAC innovatively synergizes a CLIP model with a multi-modal large language model (MLLM) to learn generalized 3D representations, where the MLLM is used for dual purposes.
First, it describes the seen category information to align with CLIP's training objective for adaptation during training. Second, it provides external hints about unknown objects complementary to visual cues during inference. 
To improve the synergy, we introduce an Additive-Bias Low-Rank adaptation (AB-LoRA), which alleviates overfitting and further enhances the generalization to unseen categories. 
With only multi-view images, DAC significantly surpasses prior arts by an average of +10.01\% mAP on four open-set 3DOR datasets. Moreover, its generalization is also validated on image-based and cross-dataset setups.
Code is available at~\href{https://github.com/wangzhichuan123/DAC}{https://github.com/wangzhichuan123/DAC}.

\end{abstract}    
\section{Introduction}
\label{sec:intro}

\begin{figure}[t]
\centering
\includegraphics[height=0.9\linewidth]{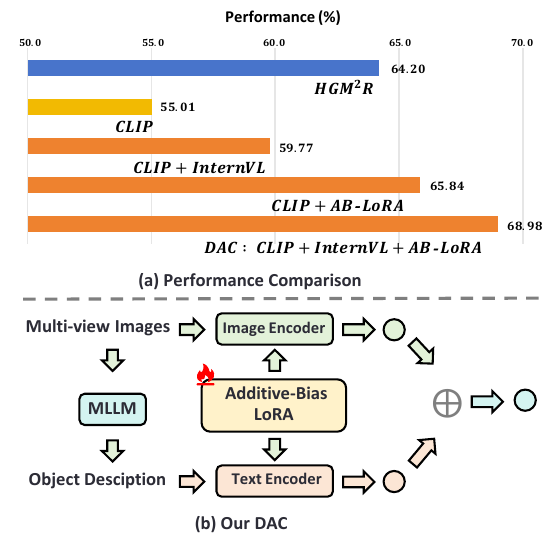}
\caption{
We empower CLIP encoders with an MLLM model (\emph{e.g.}, InternVL~\cite{chen2024internvl}) for open-set 3D object retrieval, which significantly outperforms previous state-of-the-art HGM$^{2}$R~\cite{feng2023hypergraph} (a). Unlike HGM$^{2}$R, which utilizes all modalities (point cloud, voxel, and view images) and involves test data for training, our method relies solely on multi-view inputs and eliminates test data (b). 
}
\label{fig:into_fig}
\end{figure}

Retrieving 3D objects from a given large 3D repository is an important task in 3D computer vision, whose applications include virtual reality~\cite{mangina20173d}, computer-aided design~\cite{fradi20173d}, and 3D cultural heritage preservation~\cite{koller2010research}.
Existing 3D object retrieval (3DOR) algorithms~\cite{feng2018gvcnn,wei2020view, sitzmann2019deepvoxels, achlioptas2018learning,liu2019point2sequence, nie2019mmjn, jing2021cross, wu2019multi, he2024latformer} usually learn discriminative representations on various 3D data formats (\emph{e.g.}, view images~\cite{su2015multi}, voxels~\cite{wu20153d}, and point clouds~\cite{qi2017pointnet}). Despite progress, the recent \emph{SHape REtrieval Contest (SHREC) 2022} competition~\cite{feng2022shrec} has shown that these methods have great difficulties in open-set scenarios, where the test labels are unseen during training. 
Recently, HGM$^{2}$R~\cite{feng2023hypergraph} uses all modalities and incorporates test data for training. Although improvement is achieved, it is complex and impractical in real-world applications.

The challenges of open-set 3DOR are twofold.
First, existing methods typically assume the same domain and category shared by training and testing data, resulting in low performance in open-set scenarios.
Second, these methods tend to overfit a handful of known categories due to the \emph{limited} 3D data, leading to poor generalization to unseen ones.

To tackle the issue, we are motivated by the fact that existing pre-trained large vision and language models, such as CLIP, can produce generalized and effective embeddings for a variety of data-scarce downstream tasks~\cite{huang2023clip2point,gao2024clip,sain2023clip}. 
We thus first 
establish a multi-view CLIP baseline to investigate its potential by plainly aggregating view-wise features from CLIP image encoder. 
This simple baseline empirically gives promising results, as shown in Figure~\ref{fig:into_fig}. 

But imagine how humans react when faced with a 3D object of an unknown category. Typically, we examine it from various angles to collect visual cues for inference. Yet, in addition to visual analysis, we can articulate descriptions of the object using language, thereby engaging in textual reasoning. For instance, ``a horse-like animal but with black and white stripes'', these descriptive sentences provide important attributes and greatly facilitate inference (feature matching)  of unknown categories. 
This fact encourages us to further involve textual descriptions of unknown 3D objects to visual features. 

In this paper, therefore, we introduce \textbf{D}escribe, \textbf{A}dapt, and \textbf{C}ombine (\textbf{DAC}), a novel framework empowering CLIP encoders for open-set 3DOR by combining Multi-Modal Large Language Models (MLLM). 
Specifically, inputting multi-view images, DAC consists of a three-step process:

1) \textbf{Describe}: We utilize a pretrainned MLLM model (\emph{e.g.}, InternVL~\cite{chen2024internvl}) to acquire textual information. During training, the MLLM is used to derive expressive category descriptions about the objects, which better aligns with the training objective of CLIP in adaptation. During inference time, we only use it to describe the appearance and semantic cues encoded in the multi-view images, thereby obtaining out-of-box textual knowledge for the feature combination.

2) \textbf{Adapt}: Paired multi-view images with enriched descriptions of seen categories, we introduce Low-Rank Adaptation (LoRA) in both text and visual encoders for efficient adaptation. This step effectively mitigates the domain gap between view projections and natural images with which CLIP is trained. However, we observed that LoRA always easily overfits the seen categories. We analyze it (see Sec.~\ref{sec:train_lora}) and further introduce an additive bias, which effectively enhances the generalization to unseen categories.

3) \textbf{Combine}: After CLIP adaptation, we combine both the multi-view image and textual features for shape retrieval. Thanks to the aligned feature space of CLIP for image and text inputs, we can directly embed the acquired descriptions via CLIP encoder and fuse the visual and textual cues in CLIP space via simple operations like addition.

DAC demonstrates strong performance across four existing open-set 3DOR benchmarks on five retrieval tasks. In particular, it surpasses state-of-the-art by over +10\% mAP on average in open-set 3D object retrieval.



In summary, we make the following contributions:
\begin{itemize}
 \item We propose a simple yet effective framework named DAC for open-set 3DOR. To our best knowledge, it represents the first attempt at synergizing an existing pretrained vision-language model and a multi-modal large language model to effectively address 3D data scarcity and unknown category challenges in open-set 3DOR. 
 \item To maximize its potential, we introduce a scheme for multi-view image adaptation. It makes use of enriched category descriptions to enhance text-image alignment through contrastive learning. Furthermore, we insert an additive bias to LoRA for better adaptation, effectively improving the generalization to unseen test categories.
 \item DAC performs remarkablely in open-set 3DOR with limited 3D data. In addition, it also effectively extends to various other 3DOR tasks like cross-dataset, single-image-based, and even zero-shot point cloud retrieval. 

\end{itemize}

\section{Related Work}
\label{sec:related_work}

\noindent\textbf{3D Object Retrieval.} 3D object retrieval is a long-standing task in computer vision~\cite{tangelder2004survey}.
While early research focused on hand-crafted descriptors~\cite{chen2003visual,belongie2002shape,kazhdan2003rotation}, modern works generally center around learning representations upon 3D data collections.
\emph{One group} learns based on raw 3D formats, including voxel grids~\cite{maturana2015voxnet,wu20153d,wang2017cnn}, point clouds~\cite{qi2017pointnet, qi2017pointnet++, wang2019dynamic, liu2019relation, zhao2019pointweb, guo2020deep,cheng2021net}, or meshes~\cite{feng2019meshnet,liang2022meshmae}. 
Although they can leverage 3D geometric information,
they often face computation challenges, \eg, cubic computation complexity for voxels and the irregular and unordered issue of point clouds and meshes.
\emph{Another group}~\cite{kanezaki2018rotationnet,han20193d2seqviews, esteves2019equivariant,feng2018gvcnn,dai2018siamese,he2018triplet, li2019angular,su2015multi,he2019view,wang2025teda} focuses on learning to aggregate multi-view features, exhibiting superior potential than 3D methods. 
Besides, images are 2D grid data, which is more computationally efficient. 

Despite great progress, all above methods concentrate on closed-set conditions and do not generalize well to open-set settings involving unseen 3D object categories, as per SHREC’22~\cite{feng2022shrec}. Subsequently, Feng~\etal~\cite{feng2023hypergraph} proposed jointly learning multi-modal embeddings, including multi-view images, point clouds, and voxels, and utilizing hypergraphs to connect seen and unseen category data. Although it leads to improved generalization, involving multi-modal and test set data brings extra complexity. 
In this work, we take a \emph{different} route. By leveraging current large vision and language models, we address the generalization challenges in open-set 3DOR, \emph{solely} based on multi-view images.

\begin{figure*}[ht]
\centering
\includegraphics[width=\linewidth]{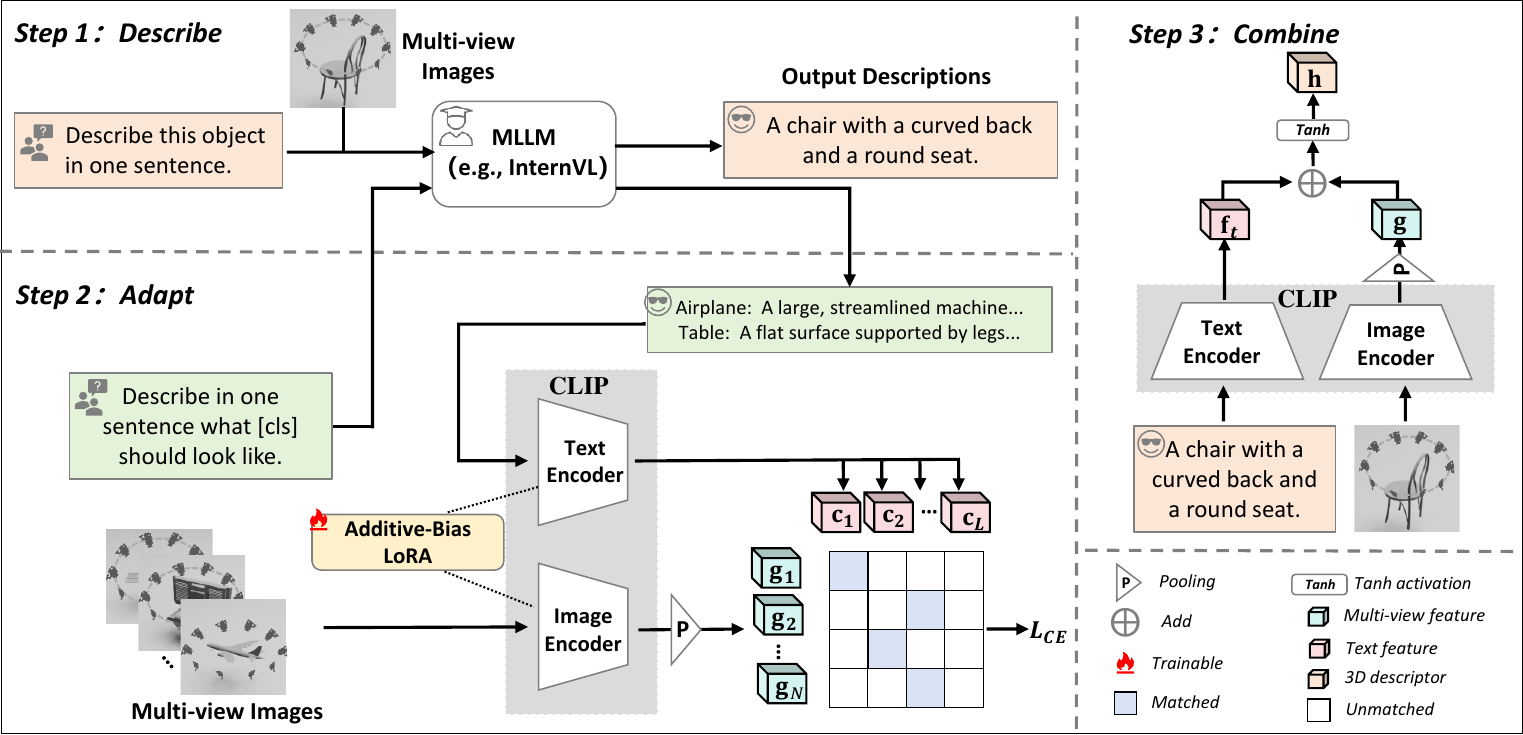}
\caption{\textbf{Overview of DAC}. Given a 3D object, we first project it into multi-view images and utilize a pretrained MLLM (\eg, InternVL) to describe the shape information. Next, we efficiently finetune a pretrained CLIP ViT model with a additive-bias LoRA to adapt to the multi-view projections. Finally, we combine the textual and multi-view embeddings to derive 3D descriptors for retrieval.}
\label{fig:framework}
\end{figure*}

\vspace*{3mm}
\noindent\textbf{Transfering Vision-Language Models in 3D.} 
Pre-trained by large-scale image-text pairs, large vision-language models (VLM) (particularly CLIP~\cite{radford2021learning}) can be utilized to produce discriminative representations for 2D depth/view images of 3D objects. 
One simple way is to use these pretrained models directly in a zero-shot manner. For instance, MV-CLIP~\cite{song2023mv} directly extracts multi-view features with pretrained CLIP's visual encoder, while works like ULIP~\cite{xue2023ulip} and ULIP-2~\cite{xue2024ulip} incorporate unified embeddings that align language, image, and point data to improve 3D shape analysis across modalities. OpenShape~\cite{liu2024openshape} builds on CLIP-based embeddings to jointly learn image, text, and 3D shape features, facilitating robust cross-modal retrieval in open-set scenarios.
Some works~\cite{zhang2022pointclip, zhu2023pointclip,huang2023clip2point} apply CLIP to process 2D depth maps of point clouds, reaching promising results. 
Even so, these methods are still restrained in performance by the amount of 3D data.

When a certain amount of downstream training data is available, it is always desirable to finetune the pretrained models to adapt to the downstream tasks for optimized results. Some works, \emph{e.g.}, CoOp~\cite{zhou2022conditional}, CLIP-Adapter~\cite{gao2024clip}, and Tip-Adapter~\cite{zhang2021tip}, learn a minuscule adapter while keeping the pretrained main model frozen.
CG3D~\cite{hegde2023clip} replaces the handcrafted prompts with learnable ones.
In this paper, we propose to use Low-Rank Adaptation (LoRA)~\cite{hu2022lora} but add a simple bias term, enabling efficiently adapts to multi-view projection images of known categories with reduced overfitting risk. In addition, it brings negligible testing latency after merging.

\noindent\textbf{Multi-model Large Language Models.} 
Multi-model Large Language Models
(MLLMs)~\cite{alayrac2022flamingo,achiam2023gpt,liu2024improved,liu2024visual,lu2023empirical,wang2024visionllm,peng2023kosmos,bai2023qwen,chen2023shikra,wang2023all,chen2024internvl} possess excellent reasoning abilities on visual content, driving progress in generic visual-linguistic tasks, such as image captioning, visual question answering (VQA), and visual dialogue. Over the past few years, many commercial and open-source MLLMs have been introduced, \emph{e.g.}, GPT-4~\cite{achiam2023gpt}, LLaVA series~\cite{liu2024improved,liu2024visual,lu2023empirical} and many others~\cite{wang2024visionllm, peng2023kosmos, bai2023qwen,chen2023shikra,wang2023all}.
In particular, InternVL~\cite{chen2024internvl} has made significant strides in enhancing the interaction between visual and language components, which is essential for tasks that require detailed object understanding.
For open-set 3D object retrieval, the ability of MLLMs to generate high-level semantic descriptions is particularly beneficial. By combining these descriptions with fine-grained visual features, our model can \emph{generalize better} to unseen categories, addressing the key challenge in open-set 3DOR.

 \section{Method}
\label{sec:method}

In open-set 3D object retrieval, our goal is to learn discriminative, more importantly, generalizable embeddings for 3D objects of unseen categories.
Formally,  let \(\mathcal{D}_{\text{train}}=\{(\mathbf{x}_i, y_i)\}_{i = 1}^{N_t}\) be a training set consisting of \(N_t\) 3D objects.
Here $\mathbf{x}_i$ represents $i$-th 3D object, $y_i$ are the corresponding known labels.
Based on it, we aim to train an embedding network \(\mathcal{E}(\cdot)\) to produce generalizable embeddings for a retrieval set \(\mathcal{D}_\text{ret}=\{(\mathbf{x}_i,\hat{y}_i)\}_{i = 1}^{N_r}\), which consists of \(N_r\) 3D objects belonging to unseen categories. The retrieval set \(\mathcal{D}_\text{ret}\) is further split into the query and target sets, denoted by \(\mathcal{D}_\text{query}\) and \(\mathcal{D}_\text{target}\).
Note that the label $\hat{y}_i$ from $\mathcal{D}_\text{ret}$  and $y_i$ from $\mathcal{D}_\text{train}$  are drawn from disjoint label sets. 

Existing works~\cite{feng2022shrec,feng2023hypergraph} deal with this task by taking advantage of multi-modal inputs of 3D objects (voxels, point clouds, multi-view images). 
By contrast, our approach simplifies the paradigm by only utilizing view images. We first project each 3D object $\mathbf{x}_i$ into a set of 2D view images from different viewpoints following the same scheme as HGM${^2}$R~\cite{feng2023hypergraph}.
We produce $M$ 2D grey-scale images $\{I_{i,m} \in \mathbb{R}^{1 \times H \times W}\}_{m=1}^M$, where $H$ and $W$ represent the height and width of the view image $I_{i,m}$, respectively. 
These multi-view projections comprehensively capture the object from various angles, ensuring the 3D structure is well-represented.
Based solely on multi-view images, we present a simple yet effective framework named \textbf{D}escribe, \textbf{A}dapt, and \textbf{C}ombine (DAC) (Figure~\ref{fig:framework}), which takes advantage of existing large multi-modal models and turn them into a strong 3D embedding learner for open-set 3D object retrieval.
DAC offers a new alternative paradigm for open-set 3D representation learning, structured as a three-step process: Describe, Adapt, and Combine. In the following, we describe each part in detail. 

\subsection{Describe: Describe 3D objects}
\label{sec:desc}

For this step, we utilize an MLLM to generate text descriptions. 
Note that any off-the-shelf MLLM can be used (see \underline{\textit{Appendix \textcolor{cyan}{I}}} for more choices). Here we mainly use InternVL~\cite{chen2024internvl} due to its superior instruction-following ability in open-ended tasks like visual question answering. We leverage the MLLM for \emph{dual purposes}.

To better align with CLIP’s contrastive pretraining objective on image-text pairs, we first use the MLLM to enrich the category with detailed descriptions. For a training set $\mathcal{D}_{\text{train}}$ with $L$ classes $\{l_i\}_{i=1}^{L}$, we use prompt: ``Describe in one sentence what [cls] should look like", replacing [cls] with each of the $L$ categories, to obtain $L$ sentences $\{t_i\}_{i=1}^{L}$. Each sentence serves as a description of the category, involving in the Adapt step to fine-tune CLIP to align with the multi-view images.
Compared with handcrafted prompts using templates ``a photo of [cls]'', using detailed descriptions is conducive to producing more generalized representations reflecting fine-grained aspects of categories.

We further enhance the inference stage by integrating MLLM-generated sentences as external hints.
Given multi-view images, we feed them into pretrained InternVL models~\cite{chen2024internvl} together with multi-view information-gathering prompt: ``There are images of an object from different angles. Describe this object in one sentence.'' The prompt instructs the models to generate descriptions $s_i$ to summarize the object’s appearances and semantic features across all views. 
They are further processed by the adapted CLIP to generate textual embeddings, further improving generalization to unseen categories.

By integrating textual guidance from the MLLM during both training and inference, we effectively combine the strengths of MLLM and CLIP for open-set 3DOR, avoiding somewhat impractical usage of test data for training~\cite{feng2023hypergraph}.  


\subsection{Adapt: Adapt CLIP to Multi-view Images}
\label{sec:train_lora}

\begin{figure}[t]
\centering
\includegraphics[width=0.9\linewidth]{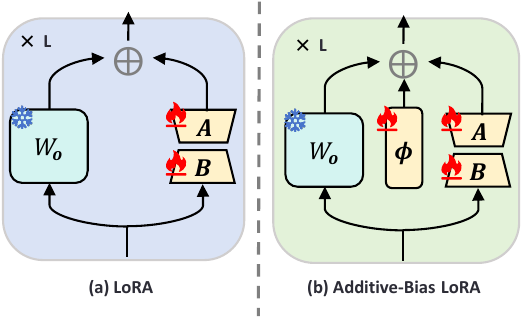}
\caption{Comparison of LoRA and Additive-Bias LoRA.}
\label{fig:lora}
\end{figure} 

The next step is to embed the multi-view images and the attached category descriptions for retrieval.
To achieve this goal, we build upon the released CLIP model~\cite{radford2021learning} which naturally supports multimodal inputs and embeds them into aligned and discriminative feature space. 

Nevertheless, CLIP is primarily pretrained on natural images, which are quite different from multi-view projections. 
To extract suitable 3D representations, it is essential to adapt to the new distributions. To this end, we first introduce Low-Rank Adaptation (LoRA)~\cite{hu2022lora} with \emph{additive bias} and insert them into the fixed CLIP model in DAC for adaption.


\noindent\textbf{A Revisit to LoRA.}
Low-Rank Adaptation (LoRA)~\cite{hu2022lora} rewrites the \emph{updates} ($\Delta\mathbf{W}$) to the pretrained weight matrix $\mathbf{W}_o \in \mathbb{R}^{d_1 \times d_2}$ of a linear layer into a product of two small low-rank matrices, denoted by $\mathbf{A} \in \mathbb{R}^{r \times d_2}$ and $\mathbf{B} \in \mathbb{R}^{d_1 \times r}$: $\mathbf{W}_o + \gamma \Delta\mathbf{W} = \mathbf{W}_o + \gamma \mathbf{BA}$. Here $\mathbf{A}$ and $\mathbf{B}$ forms a LoRA module (see Figure~\ref{fig:lora} (a)), the rank $r << \text{min}(d_1, d_2)$, and $\gamma$ is a scaling factor. 
Note that we only train $\mathbf{A}$ and $\mathbf{B}$ and keep $\mathbf{W}_o$ frozen.
$\mathbf{A}$ and $\mathbf{B}$ can also be viewed as a reparametrization and merged into $\mathbf{W}_o$ readily. 
Given the input $\mathbf{z} \in \mathbb{R}^{d_2}$, then the output $\mathbf{o} \in \mathbb{R}^{d_1}$ to a linear layer with LoRA is formulated as: 
$\mathbf{o} = \mathbf{W}_o\mathbf{z} + \gamma \Delta \mathbf{W}\mathbf{z} = \mathbf{W}_o\mathbf{z} + \gamma \mathbf{BA}\mathbf{z}$.

\noindent \textbf{Additive-bias LoRA.} For open-set 3DOR, we face a severe risk of overfitting on seen categories due to limited training data, which often causes poor generalization to testing unseen categories. 
To alleviate it, we rethink LoRA and propose to inject a ``bias'' inside, which has been widely overlooked but has unreasonable effectiveness for improving unseen category generalization.
Given loss $\mathcal{L}$, the gradient with $\Delta\mathbf{W}$ as a whole is: $\frac{\partial \mathcal{L}}{\partial \Delta\mathbf{W}} = \gamma \left( \frac{\partial \mathcal{L}}{\partial \mathbf{o}} \right) \mathbf{z}^T$.
Following the SGD update rule: $\Delta\mathbf{W} \leftarrow \Delta\mathbf{W} - \eta (\gamma \left( \frac{\partial \mathcal{L}}{\partial \mathbf{o}} \right) {\mathbf{z}}^T)$, where $\eta$ is the learning rate. 
Note that $\Delta\mathbf{W}$ directly accumulates information from $\mathbf{z}$, which comes from \textit{seen categories}. To break this tight linkage, we add a bias term $\mathbf{\Phi}\in \mathbb{R}^{d_1}$ to prevent the model from being overfitted quickly and maintain generalization to \textit{unseen categories} as much as possible:
\begin{equation}
\mathbf{o} = \mathbf{W}_o\mathbf{z} + \gamma \mathbf{BA}\mathbf{z} + \mathbf{\Phi}.
\end{equation}

To initialize, we use a standard normal distribution for $\mathbf{A}$ and set $\mathbf{B}$ and $\mathbf{\Phi}$ to zero, 
which ensures that the model starts without perturbations to the original pre-trained weights. 

In our architecture, we use this LoRA variant in both the textual and visual encoders of CLIP ViT models (see Figure~\ref{fig:lora} (b)). Following LoRA~\cite{hu2022lora}, we limit LoRA modules to the weights of the self-attention modules (\emph{i.e.}, $\mathbf{W}_{\text{query}}, \mathbf{W}_{\text{key}}, \mathbf{W}_\text{value}$). 

\noindent\textbf{Training Objective.} 
We train both textual and visual encoders in CLIP with contrastive learning, enabling better alignment between multi-view images and their associated category descriptions, with the limited 3D training data. 
Given $L$ sentences $\{t_i\}_{i=1}^{L}$ from the first step, we generate $L$ classification weights by forwarding each into the CLIP textual encoder:
\begin{equation}
\mathbf{c}_{i} = \mathcal{T}(t_i) \in \mathbb{R}^d, i=1, ..., L.
\end{equation}

Next, we feed the multi-view images $\{I_{k,m}\}_{m=1}^M$ of 3D object $\mathbf{x}_k$ into CLIP visual encoder $\mathcal{V}(\cdot)$: 
\begin{equation}
\mathbf{f}_{k,m} = \mathcal{V}(I_{k,m}) \in \mathbb{R}^d,
\end{equation}
This will give us $M$ view-wise representations $\{\mathbf{f}_{k,m}\}_{m=1}^{M}$. 
We further aggregate them into a global compact descriptor for the 3D object via mean-pooling: 
\begin{equation}
\mathbf{g}_{k} = \frac{1}{M} \sum_{m=1}^M \mathbf{f}_{k,m}.
\end{equation}

Finally, we apply cross-entropy loss to supervise the training process of LoRA, ensuring that the global features $\{\mathbf{g}_k\}$ are aligned with their corresponding text descriptions $\{\mathbf{c}_y\}$ closely in the embedding space: 
\begin{equation}
\mathcal{L}_{CE} = - \frac{1}{N_t}\sum_{k=1}^{N_t}\log \frac{\exp(\mathbf{g}_{k} \cdot \mathbf{c}_{y}/\tau)}{\sum_{i=1}^{L} \exp(\mathbf{g}_{k} \cdot \mathbf{c}_{i}/\tau)}.
\end{equation}
where $\tau$ is the temperature, and $y$ is the ground-truth category. 
By efficiently fine-tuning additive bias LoRA with limited 3D training objects, we adapt pre-trained CLIP ViT models to multi-view projections with minimal effort, while demonstrating excellent performance. 

\subsection{Combine: Combine Textual-Visual Features}
\label{sec:combine}

After the processes of description and adaptation, we proceed to combine textual and visual embeddings to form discriminative and generalized descriptors for 3D objects. 
For this, we utilize the finetuned CLIP model in Sec.~\ref{sec:train_lora} to perform feature extraction for the two modalities (\ie, descriptive text and multi-view images). 

With the generated description $s$ and multi-view images $\{I_m\}$ of 3D objects, we easily derive each modality's global representations by forwarding them to the adapted CLIP ViT models: $\mathbf{f}_{\text{t}} = \mathcal{T}(s)$, $\mathbf{f}_{m} = \mathcal{T}(I_{m})$.
Here we aggregate the multi-view representations into $\mathbf{g}$ with a simple mean-pooling layer. 
Then, we propose a weighted fusion scheme, which balances visual and textual features with a scalar and then normalizes to enhance discriminativeness:
\begin{equation}
   \mathbf{h} =  tanh (\mathbf{g} + \alpha \mathbf{f}_{t} ).
\end{equation}
where $\alpha \in [0,1]$ is a weighting factor. The $tanh$ activation function is applied for normalization after the fusion of \(\mathbf{g}\) and \(\mathbf{f}_{t}\). Finally, \(\mathbf{h} \in \mathbb{R}^d\) represents the final descriptor for the 3D object.
Without bells and whistles, this simple fusion method gives us surprisingly good performance. For retrieval, we adopt cosine similarity as the metric function. 


\subsection{Discussions} 
\noindent\textbf{Why not directly use InternVL for retrieval?}
While InternVL excels at multi-modal generative tasks, its pretraining is not specifically geared toward producing discriminative features suitable for retrieval.
In contrast, CLIP’s encoder is designed to produce more discriminative features, which are crucial for distinguishing between categories in open-set scenarios.
Empirical evidence from OS-ESB-core shows that using InternVL's embeddings led to an inadequate 38.20\% mAP, much lower than CLIP's 53.93\% mAP.


\noindent\textbf{Extending to other 3DOR.} DAC indeed represents a general framework and can be easily extended to a variety of 3DOR problems such as cross-domain retrieval. 
By leveraging semantic descriptions from an MLLM and fine-tuning discriminative models like CLIP, DAC can even handle scenarios where domain shifts present. This simple synergy showcases strong performance with minimal complexity, making it a versatile solution for various 3D retrieval tasks. 
\section{Experiments}
\label{sec:exp}
\subsection{Experimental Setup}
\label{sec:setup}

\begin{table*}[h]
\centering
\small 
\setlength{\tabcolsep}{1.0pt}
\resizebox{1.0\linewidth}{!}{
\begin{tabular}{l@{\hspace{1pt}}cccc@{\hspace{10pt}}ccc@{\hspace{10pt}}ccc@{\hspace{10pt}}ccc}
\toprule
\multirow{2}[2]{*}{Method} & \multirow{2}[1]{*}{Modality} & \multicolumn{3}{c}{OS-ESB-core} & \multicolumn{3}{c}{OS-NTU-core} & \multicolumn{3}{c}{OS-MN40-core} & \multicolumn{3}{c}{OS-ABO-core} \\
\cmidrule(lr){3-5} \cmidrule(lr){6-8} \cmidrule(lr){9-11} \cmidrule(lr){12-14}
 & & \footnotesize{mAP$\uparrow$} & \footnotesize{NDCG$\uparrow$} & \footnotesize{ANMRR$\downarrow$} & \footnotesize{mAP$\uparrow$} & \footnotesize{NDCG$\uparrow$} & \footnotesize{ANMRR$\downarrow$} & \footnotesize{mAP$\uparrow$} & \footnotesize{NDCG$\uparrow$} & \footnotesize{ANMRR$\downarrow$} & \footnotesize{mAP$\uparrow$} & \footnotesize{NDCG$\uparrow$} & \footnotesize{ANMRR$\downarrow$} \\
\midrule
\textbf{~~{Zero-shot Setup}} \\ 
OpenShape (\footnotesize{PB-CLIP B/32})~\cite{liu2024openshape} & P. & 37.64 & 18.38 & 65.57 & 25.53 & 15.41 & 74.51 & 30.31 & 46.34 & 67.01 & 40.29 & 46.09 & 59.42 \\
OpenShape (\footnotesize{PB-CLIP L/14})~\cite{liu2024openshape} & P. & 38.58 & 18.81 & 65.10 & 24.71 & 15.02 & 75.18 & 29.64 & 44.79 & 67.64 & 38.65 & 45.56 & 60.90 \\
ULIP-2 (\footnotesize{PB-CLIP G/14})~\cite{xue2024ulip} & P. & 45.14 & 21.00 & 59.15 & 31.50 & 17.89 & 68.99 & 32.76 & 48.92 & 65.22 & 44.26 & 49.04 & 55.64 \\
Uni3D (\footnotesize{Uni3D-Giant})~\cite{zhou2023uni3d} & P. & 44.42 & 20.92 & 59.96 & 32.02 & 18.04 & 68.49 & 33.21 & 50.51 & 65.11 & 45.92 & 49.79 & 53.96 \\
\textcolor{gray}{MV-CLIP$^{\ddagger}$} (\footnotesize{CLIP B/32})~\cite{song2023mv} & \textcolor{gray}{I.} & \textcolor{gray}{46.74} & \textcolor{gray}{21.72} & \textcolor{gray}{57.34} & \textcolor{gray}{47.78} & \textcolor{gray}{23.69} & \textcolor{gray}{54.48} & \textcolor{gray}{52.46} & \textcolor{gray}{66.24} & \textcolor{gray}{48.33} & \textcolor{gray}{52.55} & \textcolor{gray}{54.95} & \textcolor{gray}{48.33} \\
\textcolor{gray}{MV-CLIP$^{\ddagger}$} (\footnotesize{CLIP L/14})~\cite{song2023mv} & \textcolor{gray}{I.} & \textcolor{gray}{49.81} & \textcolor{gray}{22.75} & \textcolor{gray}{53.71} & \textcolor{gray}{\underline{57.71}} & \textcolor{gray}{\underline{26.46}} & \textcolor{gray}{\underline{45.25}} & \textcolor{gray}{\textbf{63.74}} & \textcolor{gray}{\textbf{74.85}} & \textcolor{gray}{\textbf{37.80}} & \textcolor{gray}{\underline{63.07}} & \textcolor{gray}{\underline{59.18}} & \textcolor{gray}{\underline{38.67}} \\
\cc \textbf{Our DAC} \cc (\textbf{CLIP B/32})  & \cc I. & \cc \underline{56.16} & \cc \underline{23.58} & \cc \underline{48.39} & \cc 55.03 & \cc 25.70 & \cc 48.32 & \cc 55.39 & \cc 68.08 & \cc 45.96 & \cc 60.77 & \cc 56.28 & \cc 41.44 \\
\cc \textbf{Our DAC} \cc (\textbf{CLIP L/14})  & \cc I. & \cc \textbf{56.60} & \cc \textbf{23.94} & \cc \textbf{47.61} & \cc \textbf{61.33} & \cc \textbf{27.53} & \cc \textbf{41.96} & \cc \underline{59.77} & \cc \underline{72.08} & \cc \underline{41.46} & \cc \textbf{65.93} & \cc \textbf{59.04} & \cc \textbf{36.60} \\
\midrule 
\textbf{{~~Open-set Setup}} \\ 
TCL~\cite{he2018triplet} & P., I., V.  
& 49.31 & 21.89 & 52.68 & 39.37 & 21.23 & 61.00 & 48.11 & 63.83 & 52.30 & 49.33 & 53.86 & 51.05 \\
SDML~\cite{hu2019scalable} & P., I., V.  
& 49.59 & 21.75 & 52.36 & 40.16 & 21.52 & 60.49 & 50.75 & 65.70 & 50.22 & 47.44 & 52.79 & 52.42 \\
CMCL~\cite{jing2021cross} & P., I., V.   
& 50.01 & 21.97 & 53.06 & 41.08 & 21.72 & 59.43 & 51.38 & 65.98 & 49.75 & 49.83 & 50.89 & 50.24 \\
MMSAE~\cite{wu2019multi} & P., I., V. 
& 49.88 & 22.06 & 53.69 & 40.85 & 21.70 & 59.99 & 52.08 & 66.57 & 49.00 & 50.51 & 53.80 & 50.49 \\
MCWSA~\cite{zheng2022multi} & P., I., V. 
& 49.48 & 21.34 & 53.75 & 39.22 & 20.69 & 62.14 & 48.78 & 63.85 & 51.95 & 45.61 & 51.05 & 54.70 \\
PROSER~\cite{zhou2021learning} & P., I., V. 
& 48.69 & 21.13 & 53.95 & 39.47 & 21.24 & 60.96 & 49.00 & 64.54 & 51.66 & 50.33 & 53.27 & 50.34 \\
InfoNCE~\cite{oord2018representation}  & P., I., V. 
& 50.26 & 21.91 & 52.63 & 40.03 & 21.19 & 61.09 & 47.37 & 63.31 & 53.02 & 46.83 & 52.14 & 53.50 \\
HGM$^{2}$R~\cite{feng2023hypergraph} & P., I., V. 
& 51.74 & 22.73 & 51.28 & 44.88 & 22.81 & 56.67 & \underline{64.20} & \underline{72.91} & \underline{38.27} & 63.39 & 57.96 & 37.96 \\

\cc \textbf{Our DAC} \cc (\textbf{CLIP B/32})  & \cc I. & \cc \textbf{58.70} & \cc \textbf{24.27} & \cc \textbf{45.67} & \cc \underline{59.21} & \cc \underline{27.06} & \cc \underline{44.58} & \cc 62.40 & \cc 72.63 & \cc 39.82 & \cc \underline{66.10} & \cc \underline{59.01} & \cc \underline{36.12} \\

\cc \textbf{Our DAC} \cc (\textbf{CLIP L/14})  & \cc I. & \cc \underline{57.80} & \cc \underline{24.36} & \cc \underline{47.44} & \cc \textbf{65.83} & \cc \textbf{28.78} & \cc \textbf{37.46} & \cc \textbf{68.98} & \cc \textbf{77.59} & \cc \textbf{33.87} & \cc \textbf{70.74} & \cc \textbf{60.87} & \cc \textbf{32.14} \\
\bottomrule
\end{tabular}
}

\caption{\textbf{Performance comparisons (\%) on open-set 3DOR benchmarks}. 
\textbf{Bold} and \underline{underline} indicate the best and second best results, respectively. $^{\ddagger}$ in gray means we provide \emph{ground-truth} category sets for the view selection process in MV-CLIP, which is impractical, as category information for unseen classes is unknown in open-set scenarios. 
PB-CLIP refers to PointBERT-CLIP, while P., I., and V. stand for Point Cloud, Multi-view Images, and Voxel, respectively.
}
\label{tab:main_results}
\end{table*}

\noindent{\textbf{Datasets and Evaluation Metrics}.}
We conduct extensive experiments on four public open-set 3DOR datasets~\cite{feng2023hypergraph}: OS-ESB-core, OS-NTU-core, OS-MN40-core, and OS-ABO-core. 
Each dataset is divided into training, probe, and gallery sets. 
The training set consists of seen categories for model training, while the probe and gallery sets contain unseen categories for evaluation. 
Following HGM$^{2}$R~\cite{feng2023hypergraph}, we adopt mean Average Precision (mAP), Normalized Discounted Cumulative Gain (NDCG) and Average Normalized Modified Retrieval Rank (ANMRR) metrics to report the retrieval performance. For both mAP and NDCG, higher values are better; while for ANMRR, lower is better. 

\noindent\textbf{Implementation Details.}
For a fair comparison, we follow the same rendering scheme as HGM$^2$R~\cite{feng2023hypergraph}
and project 24 view images of size $256 \times 256$ for each 3D object, for all the experiments. 
We experiment on pretrained CLIP models~\cite{radford2021learning} with ViT-B/32 and ViT-L/14 as the backbone, respectively. 
Empirically, we set the rank $r$ of AB-LoRA to 8 by default. We also regularize its input by a dropout layer with $p=0.25$. To train, we utilize Stochastic Gradient Descent (SGD) with a learning rate of $2 \times 10^{-4}$, a batch size of 4, and a cosine learning rate scheduler. The model is trained for 30 epochs on two NVIDIA RTX 4090 GPUs. 
For the MLLM, we primarily use InternVL-4B~\cite{chen2024internvl}.
For more dataset and implementation details, please refer to \textit{\underline{Appendix \textcolor{cyan}{A}}}.

\subsection{Intra-dataset Open-set 3DOR}

\noindent\textbf{Compared Methods.}
For comprehensive comparisons, we select twelve representative methods, which can be categorized into four groups: 1) Vision-language-based methods, \emph{i.e.}, Uni3D~\cite{zhou2023uni3d}, ULIP-2~\cite{xue2024ulip} and OpenShape~\cite{liu2024openshape}, which leverage CLIP for 3D learning~\footnote{We directly evaluate released models from them in a zero-shot manner. Yet, we observe worse results after fine-tuning in the open-set setup.}. 
We also re-implement a superior zero-shot method~MV-CLIP~\cite{song2023mv}, which requires a ground-truth set for view selection. 
2) Previous excellent 3DOR methods including TCL~\cite{he2018triplet},  
SDML~\cite{hu2019scalable}, and CMCL~\cite{jing2021cross}). They have shown strong performance in close-set settings and are now extended to this challenging setup. 3) Auto-encoder-based methods (MMSAE~\cite{wu2019multi}, MCWSA~\cite{zheng2022multi}), which aim to learn compressed representations in an unsupervised manner. 
4) Specially-designed open-set methods (PROSER~\cite{zhou2021learning}, InfoNCE~\cite{oord2018representation} and HGM$^{2}$R~\cite{feng2023hypergraph}), 
which is specifically crafted to handle open-set scenarios. 


\begin{table*}[h]
    \centering
    \small
    \setlength
    \tabcolsep{10pt}
\resizebox{\textwidth}{!}{%
    \begin{tabular}{ccccccc}
        \toprule
        Backbone & InternVL & AB-LoRA & OS-ESB-core & OS-NTU-core & OS-MN40-core & OS-ABO-core \\
        \midrule
        \multirow{4}{*}{CLIP ViT-B/32} 
        & \xmark & \xmark & 53.93 / 23.00 / 49.70 & 49.35 / 23.70 / 53.03 & 49.60 / 65.71 / 50.88 & 47.64 / 51.55 / 52.47 \\
        & \cmark & \xmark & 56.16 / 23.58 / 48.39 & 55.03 / 25.70 / 48.32 & 55.39 / 68.08 / 45.96 & 60.77 / 56.28 / 41.44 \\
        & \xmark & \cmark & 57.45 / 23.96 / 47.13 & 54.57 / 25.80 / 49.08 & 59.35 / 71.89 / 42.72 & 56.45 / 55.91 / 45.33 \\
        & \cmark & \cmark & \textbf{58.70} / \textbf{24.27} / \textbf{45.67} & \textbf{59.21} / \textbf{27.06} / \textbf{44.58} & \textbf{62.40} / \textbf{72.63} / \textbf{39.82} & \textbf{66.10} / \textbf{59.01} / \textbf{36.12} \\
        \midrule
        \multirow{4}{*}{CLIP ViT-L/14} 
        & \xmark & \xmark & 54.68 / 23.39 / 48.67 & 57.29 / 26.24 / 45.47 & 55.01 / 70.55 / 45.72 & 57.35 / 56.13 / 44.42 \\
        & \cmark & \xmark & 56.60 / 23.94 / 47.61 & 61.33 / 27.53 / 41.96 & 59.77 / 72.08 / 41.46 & 65.93 / 59.04 / 36.60 \\
        & \xmark & \cmark & 56.72 / 23.96 / 48.18 & 62.77 / 28.03 / 40.34 & 65.84 / 76.10 / 36.65 & 62.92 / 58.48 / 38.69 \\
        & \cmark & \cmark & \textbf{57.80} / \textbf{24.36} / \textbf{47.44} & \textbf{65.83} / \textbf{28.78} / \textbf{37.46} & \textbf{68.98} / \textbf{77.59} / \textbf{33.87} & \textbf{70.74} / \textbf{60.87} / \textbf{32.14} \\
        \bottomrule
    \end{tabular}
    }
    \caption{\textbf{Effectiveness of the designed modules}. Values are presented in mAP/NDCG/ANMRR format.}
     \label{tab:ablation}
\end{table*}

\noindent\textbf{Result Analysis.}
Table~\ref{tab:main_results} compares open-set retrieval performance between our DAC and other representative methods.
We conduct experiments under two settings: Zero-shot and Open-set.
In the zero-shot setup, DAC achieved the best results across the OS-ESB-core, OS-NTU-core, and OS-ABO-core datasets, notably surpassing the previous state-of-the-art by over 6\% on the OS-ESB-core dataset. Although MV-CLIP achieved better results on the OS-MN40-core dataset, it cannot be directly applied to open-set tasks. In MV-CLIP’s view selection process, category information must be provided, which is \emph{incompatible} with open-set scenarios where category labels are inherently unknown in advance.
In open-set setup, as shown, DAC is better on OS-ESB-core which only has 98 training objects, while much worse on the remaining three larger benchmarks.
Among existing methods, HGM$^2$R is the previous state-of-the-art. 
However, DAC with ViT-B/32 and ViT-L/14 surpass it significantly by over 6.9\% and 6.0\% in mAP on OS-ESB-core, respectively.
We further observe DAC with both backbones demonstrates remarkable performance on OS-NTU-core, bringing over 14\% and 20\% higher mAP than  HGM$^2$R, respectively.
OS-MN40-core and OS-ABO-core have larger training sets. 
With more 3D objects for training, HGM$^2$R exhibits promising outcomes by connecting seen and unseen categories with hypergraphs. 
In contrast, DAC with ViT-B/32 shows inferior results on OS-MN40-core and comparable performance on OS-ABO-core. It is noteworthy that, Unlike HGM$^2$R, we do not 1) utilize test data for training and 2) solely make use of multi-view images rather than multi-modality (voxels, point clouds, multi-view images). 
DAC with stronger ViT-L/14 further enhances the performance, surpassing HGM$^2$R greatly.


\subsection{Analyses}
In this section, we study the core design choices of DAC.
More ablation studies are induced in \textit{\underline{Appendix \textcolor{cyan}{B}}} .

\noindent\textbf{Impact of InternVL.}
We first thoroughly assess the impact of InternVL~\cite{chen2024internvl} (see \underline{\textit{Appendix \textcolor{cyan}{I}}} for more choices) on all four datasets with both CLIP ViT-B/32 and CLIP ViT-L/14. 
The multi-view features extracted by CLIP after AB-LoRA fine-tuning (DAC w.o InternVL) serve as our baselines.
As shown in Table~\ref{tab:ablation}, incorporating InternVL predictions for text embeddings leads to consistent improvements across all datasets, regardless of backbones.
As shown, on the challenging real-world OS-ABO-core dataset, we siginantly improve mAP and ANMRR by +9.65\% and +9.21\%, respectively, when adopting CLIP ViT-B/32 as the backbone. 
Similar encouraging observations are shown on OS-MN40-core and OS-NTU-core, regardless of the backbones. 
We also note that on OS-ESB-core, incorporating InternVL results in less pronounced improvements. For instance, when adopting CLIP ViT-B/32, we increase mAP from 57.45 to 58.70. It can be attributed to the nature of OS-ESB-core, which contains high-genus objects~\cite{jayanti2006developing}, such as mechanical parts, posing great difficulties for InternVL in producing semantic descriptions.
The above experiments show that incorporating InternVL can offer an effective solution to enhance DAC for open-set 3DOR. 

\noindent\textbf{Impact of Adaptation.}
AB-LoRA enables efficient adaptation of CLIP to the multi-view projection image distribution.  
Table~\ref{tab:ablation} studies the impact of adaptation on all four datasets. 
As shown, incorporating it yields significant performance gains, particularly on OS-MN40-core, which shows +7.01\% mAP improvement using ViT-B/32 and +9.21\% using ViT-L/14.
DAC incorporates a variant of LoRA for efficient adaptation but introduces negligible testing costs. 
These excellent results shows robustness, generalization, and versatility of our DAC.

\noindent\textbf{AB-LoRA \emph{v.s} LoRA.}
Open-set 3DOR requires learning robust representations from known categories in the training set to achieve generalization to unknown categories, making it highly susceptible to overfitting on known classes. To mitigate this, we introduce bias into LoRA during training. 
We conduct experiments on OS-MN40-core to evaluate the impact of introducing bias. As shown in Table~\ref{tab:ablation-lora-bias}, compared with LoRA without bias, AB-LoRA increases the performance by +2.55\% mAP. 
This empirical evidence validates the surprising effectiveness of a simple bias in enhancing the model's generalization to unknown categories.
\begin{table}[ht]
    \centering

    \small 
    \setlength{\tabcolsep}{12pt}
    \resizebox{1.0\linewidth}{!}{
    \begin{tabular}{lccc}
    \toprule
    Method & mAP$\uparrow$ & NDCG$\uparrow$ & ANMRR$\downarrow$ \\
    \midrule
    Without LoRA & 55.39 & 68.08 & 45.96 \\
    LoRA  & 59.85 & 70.25 & 41.75 \\
    \cc AB-LoRA &  \cc \textbf{62.40} &  \cc \textbf{72.63} &  \cc \textbf{39.82} \\
    \bottomrule
    \end{tabular}}
\caption{\textbf{Ablations of Additive Bias LoRA}.}
\label{tab:ablation-lora-bias}

\end{table}


\noindent\textbf{Analysis on Fusion Scheme.} We analyze two common parameterless fusion schemes for integrating textual and multi-view features: concatenation (\emph{Concat.}) and element-wise addition (\emph{Add.}). As shown in Table~\ref{tab:ablation-fusion}, the \emph{Add.} method consistently outperforms the \emph{Concat.} method across all metrics and backbones by a large margin.
For the ViT-B/32 backbone, \emph{Add.} improves mAP from 55.30 to 62.40. Likewise, for the ViT-L/14 backbone, \emph{Add.} achieves the best results with a mAP of 68.98, NDCG of 77.59, and ANMRR of 33.87. It implies that element-wise addition can more effectively utilize the two modalities to generate more discriminative representations for retrieval.
\begin{table}[ht]
    \centering
    \setlength{\tabcolsep}{10pt}
    \resizebox{1.0\linewidth}{!}{
    \begin{tabular}{lccccc}
    \toprule
    Backbone & Fusion Method & mAP$\uparrow$ & NDCG$\uparrow$ & ANMRR$\downarrow$ \\
    \midrule
    \multirow{2}{*}{ViT-B/32} 
        &  \emph{Concat.} & 55.30 & 65.12 & 46.11 \\
        & \emph{Add.} & \cc \textbf{62.40} & \cc \textbf{72.63} & \cc \textbf{39.82} \\
    \midrule
    \multirow{2}{*}{ViT-L/14} 
        & \emph{Concat.} & 56.74 & 66.08 & 45.04 \\
        &\emph{Add.} & \cc \textbf{68.98} & \cc \textbf{77.59} & \cc \textbf{33.87} \\
    \bottomrule
    \end{tabular}}
    \caption{\textbf{Impact of fusion Schemes} on different backbones.}
\label{tab:ablation-fusion}
\end{table}


\subsection{Cross-dataset Open-set 3DOR}
\noindent\textbf{Setup}. 
To evaluate DAC on cross-dataset generalization capacity, We train all the models on OS-MN40-core and evaluate them on OS-ABO-core. 

\noindent\textbf{Results}. The results are summarized in Table~\ref{tab:cross-dataset}. 
It can be observed that DAC outperforms the other compared methods impressively. 
Specifically, with the ViT-L/14 backbone, our approach surpasses the previous state-of-the-art by +12.31\% mAP. This substantial improvement highlights the effectiveness of our method in cross-dataset scenarios, showcasing its strong ability to process instances of unknown categories from an entirely distinct domain.
\begin{table}[ht]
\centering
\small
\setlength{\tabcolsep}{14pt}
\resizebox{1.0\linewidth}{!}{
\begin{tabular}{lcccc}
\toprule
\multirow{2}{*}{Method} 
& \multicolumn{3}{c}{OS-MN40-core $\rightarrow$ OS-ABO-core} \\
  & mAP$\uparrow$ & NDCG$\uparrow$ & ANMRR$\downarrow$  \\
\midrule
CMCL~\cite{jing2021cross}   & 53.90 & 53.79 & 47.28  \\
MMSAE~\cite{wu2019multi}  & 52.83 & 52.80 & 48.02  \\
MCWSA~\cite{zheng2022multi}  & 49.20 & 50.99 & 51.11 \\
PROSER~\cite{zhou2021learning} & 50.80 & 52.37 & 49.73  \\
InfoNCE~\cite{oord2018representation}& 51.63 & 52.75 & 49.16 \\
HGM$^{2}$R~\cite{feng2023hypergraph}& 57.55 & 54.14 & 45.35 \\
\cc Ours (ViT-B/32)& \cc \textbf{63.45} & \cc \textbf{57.34} & \cc \textbf{38.48}   \\
\cc Ours (ViT-L/14) & \cc \textbf{69.86} & \cc \textbf{60.13} & \cc \textbf{32.42}   \\
\bottomrule
\end{tabular}
}
\caption{\textbf{Comparisons (\%) on cross-dataset retrieval}.}
\label{tab:cross-dataset}
\end{table}

\subsection{Single Image based Open-set 3DOR}
\noindent\textbf{Setup}. Recall that OS-MN40-core contains synthetic 3D objects, whereas OS-ABO-core includes real-world 3D objects, each with an attached real-world image.
Here we only utilize one projected image (\emph{Prj.}) or one real image (\emph{Rel.}) depicting each 3D object. 
We alternate the two datasets for training and retrieval, yielding two setups: 
\emph{MN40} $\rightarrow$ \emph{ABO} and \emph{ABO} $\rightarrow$ \emph{MN40}.
In each setup, we have two subsettings: 
\emph{Prj.} $\rightarrow$ \emph{Rel.} (or \emph{Rel.} $\rightarrow$ \emph{Prj.}) and \emph{Prj.} $\rightarrow$ \emph{Prj.}.

\noindent\textbf{Results}. The results are summarized in Table~\ref{tab:vr}. 
Encouragingly, DAC still has the best generalization ability despite the great domain gap between real and synthetic data:
We note that the results under the \emph{Prj.} $\rightarrow$ \emph{Rel.} setting are better than the \emph{Prj.} $\rightarrow$ \emph{Prj.} setting. This can be attributed to both CLIP and InternVL being pre-trained on real images, allowing it to generalize well to real-world 3DOR. 
It highlights DAC's potential in the challenging real-world 3DOR.

\begin{table}[ht]
\centering
\small
\setlength{\tabcolsep}{10pt}
\resizebox{1.0\linewidth}{!}{
\begin{tabular}{lcc|cc}
\toprule
\multirow{2}{*}{Method} 
& \multicolumn{2}{c|}{\emph{MN40} $\rightarrow$ \emph{ABO}}
& \multicolumn{2}{c}{\emph{ABO} $\rightarrow$ \emph{MN40}} \\
 & \emph{Prj.} $\rightarrow$ \emph{Rel.} 
 & \emph{Prj.} $\rightarrow$ \emph{Prj.}  
 & \emph{Rel.} $\rightarrow$ \emph{Prj.} 
 & \emph{Prj.} $\rightarrow$ \emph{Prj.}  \\
\midrule
CMCL~\cite{jing2021cross}   & 34.92 & 45.58 & 40.41 & 41.40 \\
MMSAE~\cite{wu2019multi}  & 35.91 & 45.24 & 40.12 & 41.80 \\
MCWSA~\cite{zheng2022multi}  & 32.63 & 44.35 & 38.78 & 41.14 \\
PROSER~\cite{zhou2021learning} & 33.65 & 44.32 & 39.62 & 41.14 \\
InfoNCE~\cite{oord2018representation}& 33.48 & 44.22 & 39.63 & 41.12 \\
HGM$^{2}$R~\cite{feng2023hypergraph}& 43.59 & 50.68 & 48.93 & 49.77 \\
\cc Ours (ViT-B/32)& \cc \textbf{55.59} & \cc \textbf{55.00} & \cc \textbf{49.34} & \cc \textbf{49.71}  \\
\cc Ours (ViT-L/14)& \cc \textbf{63.96} & \cc \textbf{60.12} & \cc \textbf{52.92} & \cc \textbf{56.24}  \\
\bottomrule
\end{tabular}
}
\caption{\textbf{Comparisons (mAP) on single image based open-set 3DOR}. \emph{Prj.} denotes Projected Image, and \emph{Rel.} denotes Real Image. A $\rightarrow$ B denotes adopting A for training and B for retrieval.
}
\label{tab:vr}
\end{table}

\subsection{Visualization and Discussions on Limitations} 
\begin{figure}[!ht]
\centering
\includegraphics[width=\linewidth]{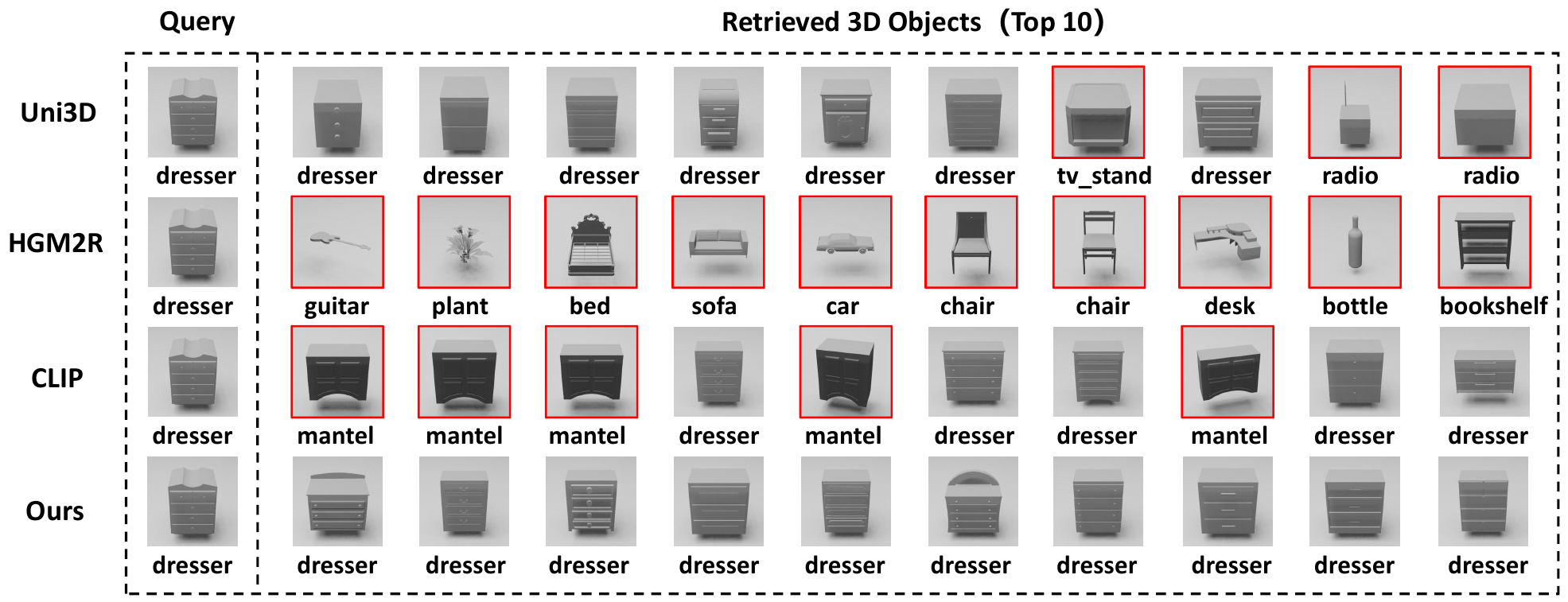}
\caption{\textbf{Retrieval example comparisons} with other methods on OS-MN40-core. Incorrect matches are in red boxes.}
\label{fig:visual_analysis}
\end{figure}

Figure~\ref{fig:visual_analysis} presents some retrieval examples. Our method can faithfully retrieve related 3D assets for the query.
Nevertheless, DAC has the following possible limitations. 1) Incorporating MLLM at inference increases costs. \emph{Yet, note that DAC already outperforms prior art HGM$^2$R even without it during inference} (see Table \ref{tab:ablation}), \emph{e.g.}, 65.84 \emph{v.s.} 63.74 mAP on OS-MN40-core. We plan to explore more training strategies like providing region-level descriptions with MLLM to enhance the training and remove the reliance. 2) DAC only fuses global textual and visual embeddings, which fails in some cases when the query and sample have similar global shapes but from hard category pair (\emph{e.g.}, wardrobe and bookshelf) (see \underline{\textit{Appendix \textcolor{cyan}{H}}}). We plan to adopt more fine-grained discriminative features to address this issue.

\section{Conclusion}
\label{sec:conclusion}

In this paper, we have presented a simple yet effective framework named DAC for open-set 3D object retrieval. 
In contrast to previous methods, we make the first attempt to synergize generative and discriminative large VLLMs to derive discriminative and generalized embeddings for open-set 3D object retrieval.
We also showed that adapting CLIP with LoRA based on multi-view images can greatly boost performance. 
Finally, we extended it to other challenging 3D representation learning tasks such as cross-dataset and zero-shot retrieval, further validating its strong generality. 

\section*{Acknowledgment}
This work is supported by National Natural Science Foundation of China (No.62302188, 62225603, 62276111 and 62076041); Fundamental Research Funds for the Central Universities (No.2662023XXQD001).

\label{sec:acknowledgment}

{
    \small
    \bibliographystyle{ieeenat_fullname}
    \bibliography{main}
}

\clearpage
\maketitlesupplementary

\appendix

Sec.~\ref{sec:details} provides more implementation details.
Sec.~\ref{sec:more_ablation} presents additional ablation studies on DAC. 
Sec.~\ref{sec:prompt} discusses InternVL prompt choices and their effects. 
Sec.~\ref{sec:pc} demonstrates the applicability of DAC in scenarios where only point cloud data is available. 
Sec.~\ref{sec:qualitative} showcases more qualitative results.
We also present additional experiments on both seen and unseen categories in Sec.~\ref{sec:seen_unseen}.
Sec.~\ref{sec:views} provides more comparisons with the state-of-the-arts by varying view numbers.
Sec.~\ref{sec:vis} give some retrieval examples to gain some insights about the limitation of our framework.
Finally, Sec.~\ref{sec:mllm} investigates the impact of different multimodal large language models (MLLMs) on DAC performance.

\section{More Dataset and Implementation Details}
\label{sec:details}
The four existing open-set 3DOR datasets, which are curated by Feng \emph{et al.}~\cite{feng2023hypergraph} are described in detail below. 
\emph{1) OS-ESB-core} is created based on ESB~\cite{jayanti2006developing}, which covers CAD objects of high genus (e.g., holes and tunnels) from the mechanical engineering domain. It includes only 98 training objects from 17 seen categories, 120 probe objects and 452 gallery objects from 24 unseen categories.
\emph{2) OS-NTU-core} has 378 training objects in 13 seen classes, 270 probe and 1,271 gallery objects in 54 unseen classes. Each object is coming from NTU~\cite{chen2003visual}.
\emph{3)} OS-MN40-core is constructed from ModelNet40~\cite{wu20153d}. It has 2,821 synthetic objects from 8 seen categories for training, 160 probe objects and 9,329 gallery objects from 32 unseen categories for testing.
\emph{4)} OS-ABO-core is a challenging, large-scale, real-word dataset with the 3D objects derived from real-word household items~\cite{collins2022abo}. It contains 1,082 training objects in 4 seen categories, 85 probe objects and 5,455 gallery objects divided into 17 unseen categories. 

For fair comparisons with off-the-shelf point cloud encoders, we further extend our framework by taking depth maps from point clouds. 
For this experiment, we have curated a zero-shot ZS-Objaverse-Core based on the large-scale Objaverse dataset.
The original Objaverse~\cite{deitke2023objaverse} contains 46,832 shapes across 1,156 LVIS categories.
We further split each category of Objaverse-LVIS into a query set and a target set with a 20\%/80\% ratio, resulting in a total of 8,798 query samples and 37,407 target samples.
For the experiment, 10 depth maps are projected for each point cloud online following~\cite{zhu2023pointclip}. 

For a comprehensive comparison, we reimplement MV-CLIP~\cite{song2023mv} to evaluate its performance on open-set 3D datasets. 
For textual prompts, we use a pre-defined template: “a synthetic 3D model view of [cls] with different angles”. 
It is important to note that we must provide \emph{ground-truth category} to MV-CLIP for view selection, which is not suitable for open-set retrieval, where the category information for unseen categories is unknown.
We select $M_{\text{selec}}$ views based on entropy and perform mean pooling over them. Following MV-CLIP, 
$M_{\text{selec}}$ is set to 4. 


For ULIP-2~\cite{xue2024ulip}, we simply use the open-source PointBERT-CLIP ViT-G/14 pre-trained model as the backbone and directly employ the output features from the last layer of the model for retrieval. Similarly, For OpenShape~\cite{liu2024openshape}, we use the PointBERT-CLIP ViT-B/32 and PointBERT-CLIP ViT-L/14 models, with the extracted features for retrieval directly. It is worth mentioning that we also experimented with fine-tuning both ULIP-2 and OpenShape for open-set setups. However, fine-tuning leads to worse performance, suggesting that the irregular point clouds are somewhat fragile representations that easily overfit to known categories in open-set setups.

\section{More Ablations}
\label{sec:more_ablation}


\subsection{Impact of Rank Number} 
\label{sec:rank}
We investigate the effect of decomposed matrix rank numbers in LoRA by setting it to 2, 4, 8, and 12. As shown in Table~\ref{tab:ablation-rank}, increasing the rank from 2 to 8 leads to consistent improvements in mAP, NDCG, and ANMRR metrics. However, further increasing the rank to 12 results in a slight decline in performance. 
A rank of 8 strikes the best balance between performance and training complexity, and thus, all experiments are conducted with rank = 8.

\begin{table}[ht]
    \centering

    \setlength{\tabcolsep}{14pt}
    \resizebox{1.0\linewidth}{!}{
    \begin{tabular}{ccccc}
    \toprule
    LoRA Rank& mAP$\uparrow$ & NDCG$\uparrow$ & ANMRR$\downarrow$ \\
    \midrule
        2 & 62.00 & 72.36 & 40.25 \\
        4 & 62.25 & 72.17 & 39.91 \\
        8 & \textbf{62.40} & \textbf{72.63} & \textbf{39.82} \\
        12 & 62.17 & 72.13 & 39.92 \\
    \bottomrule
    \end{tabular}}
    \caption{Ablation of LoRA rank number with CLIP ViT-B/32 as the backbone on the OS-MN40-core dataset.}
\label{tab:ablation-rank}
\end{table}

\subsection{Impact of Fusion Weight $\alpha$} 
\label{sec:fusion_rate}

\begin{figure}[h]
\centering
\includegraphics[width=0.95\linewidth]{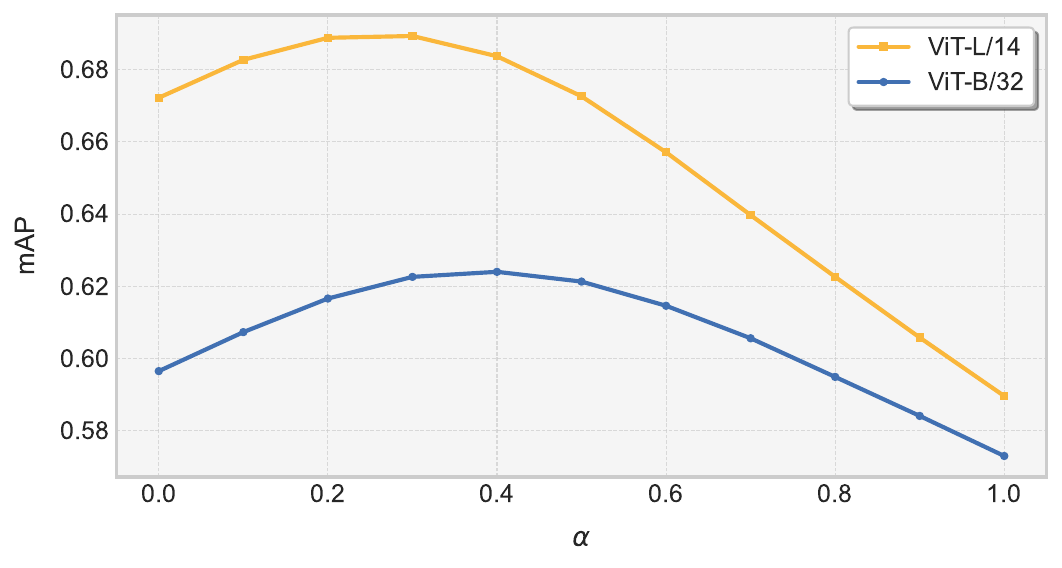}
\caption{Impact of fusion weight $\alpha$ on OS-MN40-core.}
\label{fig:fusion-alpha}
\end{figure}

The hyper-parameter $\alpha$ governs the relative weights of text features to image ones.
To study its effects, we adjust the fusion ratio $\alpha$ within the range of 0 to 1 and conduct experiments on OS-MN40-core. 
The results are summarized in Figure~\ref{fig:fusion-alpha}.
As shown, an appropriate choice of $\alpha$ is essential for good retrieval performance. For instance, with CLIP ViT-L/14, we observe the performances are gradually improved by increasing $\alpha$ from 0 to 0.25. When $\alpha$ is set to 0.25, we have the best mAP of 68.98\%.
However, further increasing $\alpha$ results in a decline in performance. 
Similar phenomena are also observed when using CLIP ViT-B/32, as well as other datasets. 
In Table~\ref{tab:fusion-alpha}, we further provide the optimal values for $\alpha$ across all the datasets. 
Interestingly, we find that on the OS-ESB-core dataset, a smaller $\alpha$ gives better results (\emph{i.e.}, 0.1). In contrast, for other datasets, especially the OS-ABO-core dataset, a larger value is preferred.
We assume that it is difficult to derive accurate text embeddings for OS-ESB-core, which consists of high-genus mechanical parts~\cite{jayanti2006developing}. In contrast, for common semantic real-world categories, text embeddings from InternVL are more accurate, and thus more weights are needed.
We set these values as our default configurations for our experiments. 

\begin{table}[ht]
    \centering
    
    \resizebox{1.0\linewidth}{!}
    {
        \begin{tabular}{cccccc}
        \toprule
        Dataset & Backbone & $\alpha$ & mAP$\uparrow$ & NDCG$\uparrow$ & ANMRR$\downarrow$ \\
        \midrule
        \multirow{2}{*}{OS-ESB-core} 
            & ViT-B/32 & $\alpha=0.1$ & 58.70 & 24.27 & 45.67 \\
            & ViT-L/14 & $\alpha=0.1$   & 57.80 & 24.36 & 47.44 \\
            \cmidrule{1-6}
        \multirow{2}{*}{OS-NTU-core} 
            & ViT-B/32 & $\alpha=0.6$   & 59.21 & 27.06 & 44.58 \\
            & ViT-L/14 & $\alpha=0.3$   & 65.83 & 28.78 & 37.46 \\
            \cmidrule{1-6}
        \multirow{2}{*}{OS-MN40-core} 
            & ViT-B/32 & $\alpha=0.4$   & 62.40 & 72.63 & 39.82 \\
            & ViT-L/14 & $\alpha=0.25$   & 68.98 & 77.59 & 33.87 \\
            \cmidrule{1-6}
        \multirow{2}{*}{OS-ABO-core} 
            & ViT-B/32 & $\alpha=0.85$   & 66.10 & 59.01 & 36.12 \\
            & ViT-L/14 & $\alpha=0.7$  & 70.74 & 60.87 & 32.14 \\
        \bottomrule
        \end{tabular}
    }

    \caption{Optimal $\alpha$ values across different datasets and backbones. }
    \label{tab:fusion-alpha}
    \vspace{-10pt} 
\end{table}

\subsection{Impact of Normalization}
\label{sec:activation}

After fusing image and text features with elementwise summation, it is beneficial to further utilize activation functions for normalization. 
We have evaluated three commonly-used activation functions: \emph{ReLU}, \emph{Tanh}, and \emph{Sigmoid}, on the OS-ABO-core dataset. 
As shown in Table~\ref{tab:ablation-activation}, with \emph{ReLU}, DAC attains inferior results. However, when adopting \emph{Sigmoid} and \emph{Tanh}, we observe consistent improvements, with \emph{Tanh} being more advantageous. Especially, for CLIP ViT-B/32, \emph{Tanh} brings an improvement of over 1\% mAP. It demonstrates that adopting \emph{Tanh} for normalization is an effective option to further augment the discriminativeness of the derived 3D descriptors. 

We also analyze why \emph{Tanh} yields better results. The \emph{Tanh} function maps input values to the range \([-1, 1]\), aligning well with the vector space used by CLIP for retrieval. This property helps the model produce evenly distributed outputs that are consistent with CLIP's embedding structure. In contrast, \emph{ReLU} sets negative values to zero, retaining only positive values, which can lead to information loss and disrupt the symmetry required for the relative relationships in CLIP’s vector space. On the other hand, \emph{Sigmoid} restricts outputs to the \([0, 1]\) range, which can weaken vector directionality and produce many low-magnitude weights, making it less compatible with CLIP's feature distribution. Thus, \emph{Tanh} emerges as the most suitable choice, preserving the necessary feature balance and enhancing the robustness of the 3D descriptors for retrieval.

\begin{table}[ht]
    \centering
    
    \small 
    \setlength{\tabcolsep}{10pt}
    \resizebox{1.0\linewidth}{!}{
    \begin{tabular}{lccccc}
    \toprule
    Backbone & Activation Function & mAP$\uparrow$ & NDCG$\uparrow$ & ANMRR$\downarrow$ \\
    \midrule
    \multirow{4}{*}{ViT-B/32} 
        & -  & 64.63 & 58.52 & 37.79 \\
        & \emph{ReLU}  & 63.13 & 58.03 & 39.08 \\
        & \emph{Sigmoid}  & 65.58 & 58.82 & 37.08 \\
        & \cc \emph{Tanh} & \cc \textbf{66.10} & \cc \textbf{59.01} & \cc \textbf{36.12} \\
    \midrule
    \multirow{4}{*}{ViT-L/14} 
        & -  & 69.33 & 60.43 & 32.98 \\
        & \emph{ReLU}  & 68.79 & 60.23 & 33.62 \\
        & \emph{Sigmoid}  & 69.95 & 60.60 & 32.83 \\
        & \cc \emph{Tanh} & \cc \textbf{70.74} & \cc \textbf{60.87} & \cc \textbf{32.14} \\
    \bottomrule
    \end{tabular}}
    \caption{Impact of Activation Functions on OS-ABO-core.}
\label{tab:ablation-activation}
\end{table}

\begin{table*}[h]
    \centering
    
    \small
    \setlength\tabcolsep{16pt}
\resizebox{\textwidth}{!}{%
    \begin{tabular}{ccccccc}
        \toprule
        Description & Method & OS-ESB-core & OS-NTU-core & OS-MN40-core & OS-ABO-core \\
        \midrule
        \multirow{2}{*}{Hand-crafted} & \emph{Images only} & 57.19 / 23.95 / 47.18 & 54.66 / 25.73 / 49.17 & 58.96 / 71.65 / 42.94 & 56.70 / 56.05 / 44.55 \\
        & \emph{Images with texts} & \cc \textbf{+1.15} / \textbf{+0.23} / \textbf{-0.76} & \cc \textbf{+3.81} / \textbf{+1.13} / \textbf{-4.24} & \cc  \textbf{+2.88} / \textbf{+0.82} / \textbf{-2.65} & \cc \textbf{+9.06} / \textbf{+2.80} / \textbf{-7.98} \\
        \midrule
        \multirow{2}{*}{Generated} & \emph{Images only} & 57.45 / 23.96 / 47.13 & 54.57 / 25.80 / 49.08 & 59.35 / 71.89 / 42.72 & 56.45 / 55.91 / 45.33 \\
         &  \emph{Images with texts} & \cc \textbf{+1.25} / \textbf{+0.31} / \textbf{-1.46} & \cc \textbf{+4.64} / \textbf{+1.26} / \textbf{-4.50} & \cc \textbf{+3.05} / \textbf{+0.74} / \textbf{-2.90} & \cc \textbf{+9.65} / \textbf{+3.10} / \textbf{-9.21} \\
        \bottomrule
    \end{tabular}
    }
    \caption{Impact of different training text descriptions with ViT-B/32 as the backbone. Values are presented in mAP/NDCG/ANMRR format.}
     \label{tab:ablation-train}
\end{table*}

\subsection{Training strategy}

We further investigate the impact of different textual descriptions during LoRA training and their influence on text fusion during retrieval. 
We have tested two settings: the first uses a fixed, hand-crafted template (``a synthetic 3D model view of [cls] with different angles"), while the second employs descriptions generated by InternVL~\cite{chen2024internvl} for each class. The experimental results are summarized in Table~\ref{tab:ablation-train}.

As shown, both training strategies yield comparable performance when retrieval is conducted using only image features. 
However, descriptions generated by InternVL demonstrate superior results in scenarios where text fusion is applied during retrieval. This improvement can be attributed to the consistency between the descriptions used during training and those generated by InternVL for text fusion in the retrieval phase. Fine-tuning with InternVL-generated descriptions enables more seamless integration of visual and textual features, enhancing the retrieval performance.

We have also experimented with providing InternVL with view projections and label information during training, allowing it to generate a description for each training sample. While this approach may appear to better align with the testing process, we have observed that the unstable descriptions generated for individual samples lead the model to focus too much on the specific characteristics of each sample, neglecting the learning of more stable, category-level features. This phenomenon is more pronounced in the OS-ESB-core dataset with fewer samples. As shown in Table~\ref{tab:ablation-train-2}, using individual-level descriptions achieved only 48.78 mAP, which is significantly lower than the 58.70 mAP obtained with category-level descriptions. Therefore, we have opted to provide a single description per category rather than per sample, as this can yield more consistent and effective results.

\begin{table}[ht]
    \centering
    
    \small
    \resizebox{1.0\linewidth}{!}
    {
        \begin{tabular}{lcccc}
        \toprule
        Dataset & Description & mAP$\uparrow$ & NDCG$\uparrow$ & ANMRR$\downarrow$ \\
        \midrule
        \multirow{2}{*}{OS-ESB-core} 
            & \emph{Individual-level} & 48.78 & 21.90 & 54.37 \\
            & \emph{Category-level} & \cc \textbf{58.70} & \cc \textbf{24.27} & \cc \textbf{45.67} \\
        \bottomrule
        \end{tabular}
    }
    \caption{Impact of different description generation methods with ViT-B/32 as the backbone.}
    \label{tab:ablation-train-2}
\end{table}

\section{More Choices for InternVL Prompts} 
\label{sec:prompt}

The selection of prompts is crucial for the responses generated by InternVL, yet finding a universal prompt that fits all scenarios is quite challenging. Using ViT-L/14 as the backbone, we have tested the impact of two different prompts:

- Prompt A: `` There are images of an object from different angles. Describe this object in one sentence." (This is the default prompt used in our method.)

- Prompt B: `` There are images of an object from different angles. Describe this object's shape information in one sentence."

Prompt A attempts to derive explicit category information, while prompt B attempts to derive descriptive shape information. 
The experimental results are detailed in Table~\ref{tab:prompt}. From the table, we can observe that in the OS-ESB-core, OS-NTU-core, and OS-MN40-core datasets, the results from both prompts are relatively similar. However, in the OS-ABO-core dataset, the effect of the prompt is more pronounced. For instance, the mAP for Prompt A is 68.40\%, whereas Prompt B yields a lower mAP of 66.43\%.

\begin{table}[ht]
    \centering
    
    \resizebox{1.0\linewidth}{!}
    {
        \begin{tabular}{lcccc}
        \toprule
        Dataset & Prompt & mAP$\uparrow$ & NDCG$\uparrow$ & ANMRR$\downarrow$ \\
        \midrule
        \multirow{2}{*}{OS-ESB-core} 
            & A & 57.80 & 24.36 & 47.44 \\
            & B & 57.89 & 24.37 & 47.31 \\
            \cmidrule{1-5}
        \multirow{2}{*}{OS-NTU-core} 
            & A & 65.83 & 28.78 & 37.46 \\
            & B & 65.58 & 28.74 & 37.59 \\
            \cmidrule{1-5}
        \multirow{2}{*}{OS-MN40-core} 
            & A & 68.98 & 77.59 & 33.87 \\
            & B & 68.81 & 77.56 & 34.18 \\
            \cmidrule{1-5}
        \multirow{2}{*}{OS-ABO-core} 
            & A & \textbf{70.74} & \textbf{60.87} & \textbf{32.14} \\
            & B & 68.68 & 59.23 & 33.57 \\
        \bottomrule
        \end{tabular}
    }
    \caption{Impact of different prompts.}
    \label{tab:prompt}
\end{table}

We randomly select a chair object from the OS-ABO-core dataset and input it into InternVL using different prompts, resulting in the following two outputs: ``A classic brown leather armchair with a high back and rounded armrests, featuring subtle nailhead trim along its edges," and ``This object has a rounded, high-backed shape with a broad seat, armrests, and a slight outward curve on the back and arms." 
We hypothesize that the statement generated by Prompt A benefits from more explicit label information, leading to superior results. Conversely, inaccurate label information could adversely affect the model's expressiveness.
Therefore, identifying a universally applicable prompt is quite challenging, and further exploration into prompt design is essential.

\noindent\textbf{Why not just let InternVL judge the categories?}
We also conduct a simple experiment to evaluate the performance of directly using InternVL for category classification. On the challenging OS-ESB-core dataset, even when provided with ground-truth category options, InternVL achieved only 11\% accuracy. Incorrect category predictions in such cases can be catastrophic for our task, as they severely disrupt feature representation and retrieval. Therefore, instead of relying on InternVL for classification, we opted to use it for generating descriptive information, which provides more robust and generalized representations.

In conclusion, while prompt selection plays a critical role, the design space for prompts remains vast, and further exploration is required to optimize their effectiveness.

\section{Extending to Point Cloud Retrieval}
\label{sec:pc}
\noindent\textbf{Setup}.
For this experiment, we curate ZS-Objaverse-Core based on Objaverse-LVIS, which is an annotated subset of Objaverse~\cite{deitke2023objaverse}, for zero-shot point cloud retrieval. 
For fair comparisons with off-the-shelf point cloud encoders OpenShape~\cite{liu2024openshape} and ULIP series~\cite{xue2023ulip,xue2024ulip}, our baseline only takes depth images from point clouds with the online projection scheme~\cite{zhu2023pointclip}. 

\noindent\textbf{Results}. As shown in Table~\ref{tab:objaverse}, our method based on depth maps also achieves superior performance, surpassing ULIP-2 by +1.22\% mAP. Note that all compared methods require huge resources to train on a large-scale 3D dataset of point cloud, text, and image triplets. By contrast, our baseline offers a much cheaper solution without 3D training.

\begin{table}[ht]
    \centering
    \small 
    \setlength{\tabcolsep}{1pt}
    \resizebox{1.0\linewidth}{!}{
    \begin{tabular}{lccccc}
    \toprule
    Method & Backbone & mAP$\uparrow$ & NDCG$\uparrow$ & ANMRR$\downarrow$ \\
    \midrule
    OpenShape (point cloud) & PointBERT-CLIP ViT-L/14 & 11.93 & 14.10 & 85.40 \\
    ULIP (point cloud) & PointMLP - SLIP   & 6.69 & 9.17 & 90.82 \\ 
    ULIP-2 (point cloud) & PointBERT - CLIP ViT-G/14  & 18.15 & 19.34 & 79.03 \\ 
    \cc \textbf{Ours (depth image)} & \cc \textbf{CLIP ViT-L/14}  & \cc \textbf{19.37} & \cc \textbf{20.15} & \cc \textbf{78.23} \\  
    \bottomrule
    \end{tabular}}
\caption{\textbf{Performance on ZS-Objaverse-Core}.}
\label{tab:objaverse}
\end{table}

\section{Qualitative Results}
\label{sec:qualitative}

\begin{figure*}[t]
\centering
\includegraphics[width=1.0\textwidth]{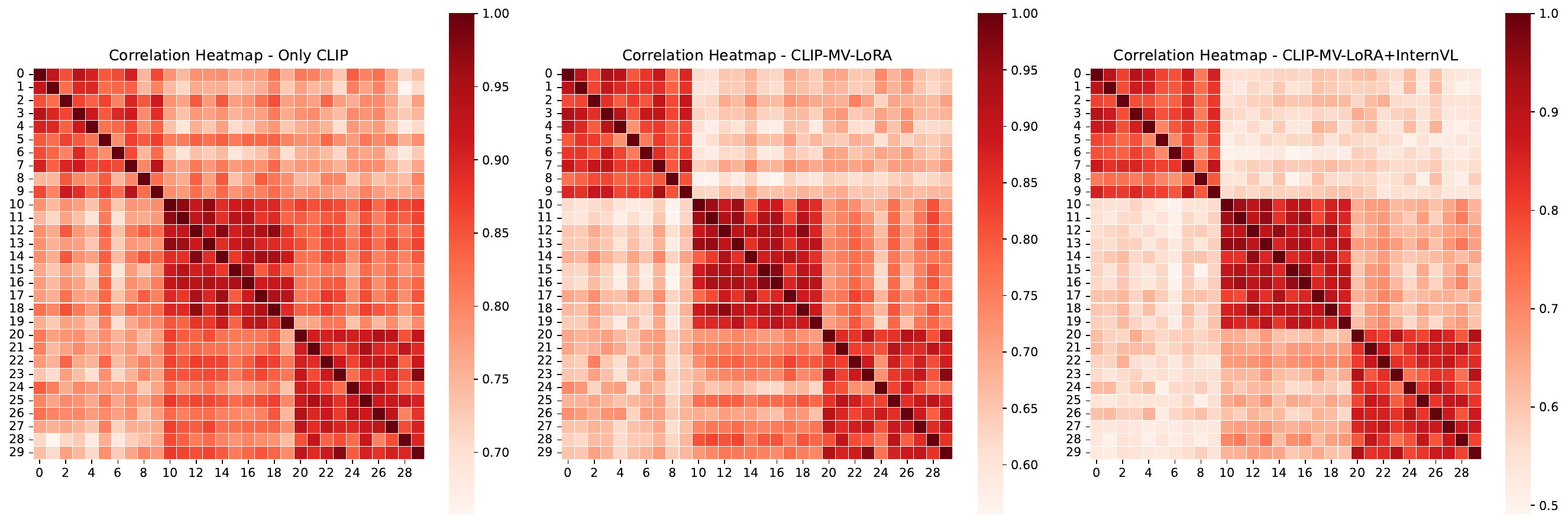}
\caption{Correlation heatmaps of features from 3 randomly selected classes, each with 10 samples on OS-ABO-core. Darker red indicates higher similarity.}
\label{fig:heatmap}
\end{figure*}

To further understand how our method improves the representations, we visualize correlation heatmaps. Specifically, we randomly select three categories from OS-ABO-core, with ten samples chosen from each category. We draw three heatmaps: the first represents features extracted from the original CLIP, the second shows features extracted using CLIP with LoRA added, and the third depicts features obtained after fusing with InternVL on top of the LoRA-enhanced features. The detailed visual results are illustrated in Figure~\ref{fig:heatmap}. As shown, we observe that, compared to the original CLIP, the addition of LoRA significantly increases the similarity among samples within the same category while reducing similarity with samples from different categories. This effect becomes even more evident in the third figure, which incorporates text features extracted by InternVL. This indicates that DAC not only enhances intra-class similarity but also effectively diminishes inter-class similarity, thereby markedly improving the discriminative power of the retrieval features.

\section{Retrieval on Seen and Unseen Categories}
\label{sec:seen_unseen}

In real-world applications, the ability to retrieve 3D objects of both seen and unseen categories is crucial. In this section, we follow HGM$^{2}$R~\cite{feng2023hypergraph} and split the ModelNet40 dataset into two subsets: \(D_{\text{S}}\) (for seen categories) and \(D_{\text{U}}\) (for unseen categories). Each subset consists of 20 categories, and the 3D objects within each category are further divided into training sets \( D_{\text{S}}^{\text{tr}} / D_{\text{U}}^{\text{tr}} \) and retrieval sets \( D_{\text{S}}^{\text{re}} / D_{\text{U}}^{\text{re}} \), with 80\% of the data used for training and 20\% for retrieval. The models are trained on \( D_{\text{S}}^{\text{tr}} \) and evaluated separately on the seen categories \( D_{\text{S}}^{\text{re}} \) and unseen categories \( D_{\text{U}}^{\text{re}} \).

As shown in Table~\ref{tab:seen_unseen}, our method has competitive performance on seen categories compared to other approaches by solely relying on multi-view images. More importantly, we achieve superior results on unseen categories, reaching an mAP of 86.27\%, surpassing previous state-of-the-art HGM$^{2}$R~\cite{feng2023hypergraph} by a large margin. Notably, our model demonstrates a reduced performance gap between seen and unseen categories, with only a 4.85\% performance difference—significantly lower than the 11.87\% gap observed with HGM$^{2}$R and the 19.73\% gap with InfoNCE. This indicates that our approach is particularly effective for retrieving unknown categories, allowing for an enhancement in unseen performance while maintaining competitive results on seen categories. Thus, our method is especially advantageous in complex environments, where retrieval performance for unseen categories is crucial.
\begin{table}[ht]
\small
\setlength{\tabcolsep}{4pt}
\resizebox{1.0\linewidth}{!}{
\centering
\begin{tabular}{lcc|cc}
\toprule
\multirow{2}{*}{Method} 
& \multicolumn{2}{c|}{On Seen Categories} 
& \multicolumn{2}{c}{On Unseen Categories} \\
 & mAP$\uparrow$ & Recall@100$\uparrow$ 
 & mAP$\uparrow$ & Recall@100$\uparrow$     \\
\midrule
TCL~\cite{he2018triplet}    & 93.50 & 82.14 & 73.92 & 71.76 \\
MMJM~\cite{nie2019mmjn}   & 91.99 & 80.78 & 73.07 & 71.38 \\
SDML~\cite{hu2019scalable}   & 88.50 & 78.50 & 74.69 & 72.39 \\
CMCL~\cite{jing2021cross}   & 90.99 & 79.60 & 75.21 & 72.49 \\
MMSAE~\cite{wu2019multi}  & 88.72 & 78.61 & 76.03 & 72.94 \\
MCWSA~\cite{zheng2022multi}  & 85.70 & 76.83 & 72.89 & 70.56 \\
PROSER~\cite{zhou2021learning} & 87.71 & 77.78 & 74.93 & 72.56 \\
InfoNCE~\cite{oord2018representation} & 93.65 & 82.19 & 73.92 & 71.64 \\
HGM$^{2}$R~\cite{feng2023hypergraph} & \textbf{94.10} & \textbf{82.47} & 82.23 & 78.21 \\
\textbf{Ours (ViT-B/32)}& 87.96 & 77.89 & \textbf{83.00} & \textbf{79.10}   \\
\textbf{Ours (ViT-L/14)}& 91.12 & 80.46 & \textbf{86.27} & \textbf{81.76}  \\
\bottomrule
\end{tabular}
}
\caption{Separate retrieval results on both seen and unseen categories.}
\label{tab:seen_unseen}
\end{table}

\begin{figure*}[ht]
\centering
\includegraphics[width=\textwidth]{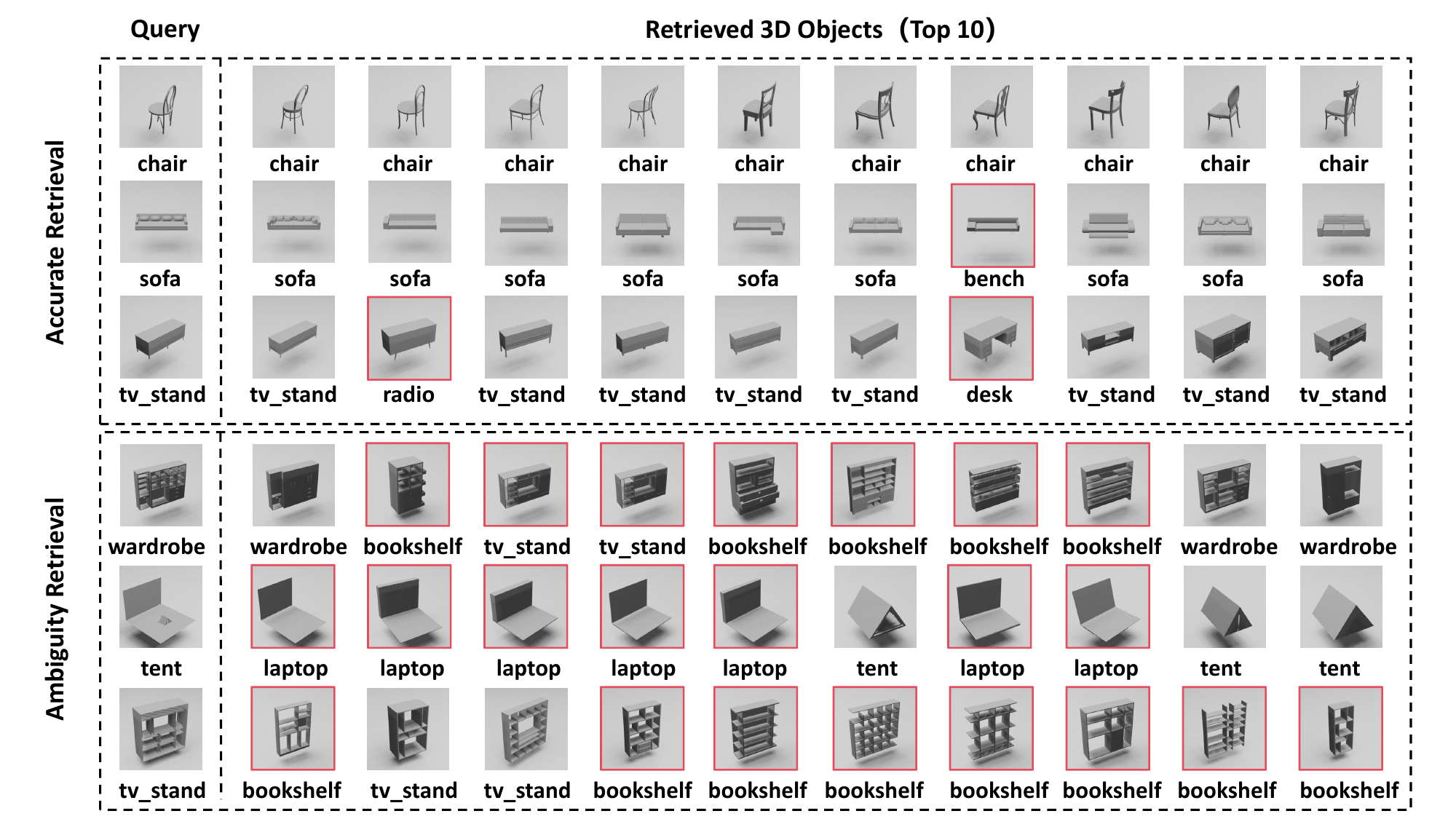}
\caption{Retrieval examples on OS-MN40-core. Incorrect matches are in red boxes.}
\label{fig:visual_analysis_2}
\end{figure*}
\section{More Experiments on View Numbers}
\label{sec:views}
\begin{figure}[h]
\centering
\includegraphics[width=0.95\linewidth]{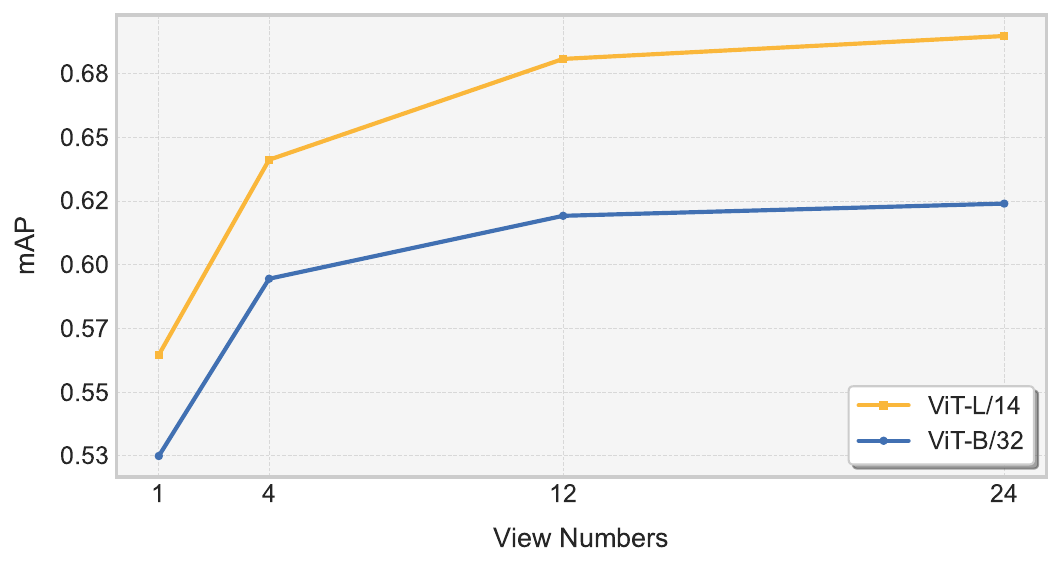}
\caption{Impact of View Numbers with different backbones.}
\label{fig:views}
\end{figure}

We further study the impact of the number of view images on the OS-MN40-core dataset.
The adjustment of view numbers affects two components: the number of views input into CLIP and the number of views input into InternVL. In our experimental setup, the view counts for both components are kept the same.
As shown in Figure~\ref{fig:views}, the mAP values increase with the number of views across different backbones. It suggests that additional views provide more detailed and accurate information about 3D objects, giving better retrieval performance. 

We also compare with other competing methods under the same number of views, as summarized in Table~\ref{tab:multi-view}. The results indicate that our approach, particularly with the ViT-L/14 backbone, outperforms other methods in both the 4-view and 12-view settings. Specifically, we achieve 64.12\% mAP and 68.08\% mAP for the ViT-L/14 backbone, respectively,  surpassing the previous best method HGM$^{2}$R greatly. 

\begin{table}[ht]
\small 
\centering
\setlength{\tabcolsep}{24pt}
\resizebox{1.0\linewidth}{!}{
\begin{tabular}{lcccc}
\toprule
\multirow{2}{*}{Method} 
 & \multicolumn{2}{c}{Number of Views} \\
  & 4-v & 12-v  \\
\midrule
TCL~\cite{he2018triplet}    & 46.46 & 47.36 \\
MMJM~\cite{nie2019mmjn}   & 46.81 & 48.11 \\
SDML~~\cite{hu2019scalable}   & 49.60 & 50.75 \\
CMCL~\cite{jing2021cross}   & 50.06 & 51.38 \\
MMSAE~\cite{wu2019multi}  & 50.85 & 52.09 \\
MCWSA~\cite{zheng2022multi}  & 47.28 & 48.78 \\
PROSER~\cite{zhou2021learning} & 48.45 & 49.00 \\
InfoNCE~\cite{oord2018representation} & 46.46 & 47.37 \\
HGM$^{2}$R~\cite{feng2023hypergraph} & 63.36 & 64.20 \\
\textbf{Ours (ViT-B/32)}& 59.45 & 61.92 \\
\textbf{Ours (ViT-L/14)}& \textbf{64.12} & \textbf{68.08} \\
\bottomrule
\end{tabular}
}
\caption{Performance comparison on view numbers.}
\label{tab:multi-view}
\end{table}

\section{More Visualization}
\label{sec:vis}
To gain more insights into our framework, we provide some retrieval examples of our method, especially including some failure cases, on OS-MN40-core. 
As shown in Figure~\ref{fig:visual_analysis_2}, for objects of easy categories (\eg, chair), our method produces discriminative 3D representations for accurate retrieval. However, it fails when two objects have similar global appearances but from distinct categories. For instance, a tent instance (row 5) globally looks like a laptop object. Yet,  notice that a laptop has ver distinct local features on the integrated keyboard. The keyboard serves as a strong discriminative cue for identifying a laptop. 
In the future, we plan to emphasizing these local features during the representation learning process, which could potentially avoid these failure cases.

\section{More Choices of MLLM}
\label{sec:mllm}
Our framework is compatible with any off-the-self pretrained MLLM, enabling seamless integration of the latest advancements in multimodal learning.
To study it, we experiment with different choices for multi-modal large language models (MLLMs). 
Table~\ref{tab:ablation-mllm} shows that DAC's performance improves progressively as InternVL~\cite{chen2024internvl} scales from 1B to 8B parameters. It suggests that MLLMs with stronger reasoning capabilities lead to better results.
Furthermore, the use of Qwen2.5-VL~\cite{Qwen2.5-VL} further enhances DAC performance, highlighting the potential of DAC.

\begin{table}[ht]
    \centering
    
     \vspace{-2pt}
    \small 
    \setlength{\tabcolsep}{12pt}
    \resizebox{1.0\linewidth}{!}{
    \begin{tabular}{lccc}
    \toprule
    MLLM & mAP$\uparrow$ & NDCG$\uparrow$ & ANMRR$\downarrow$ \\
    \midrule
    InternVL-1B~\cite{chen2024internvl} & 61.48 & 72.64 & 40.60 \\
    InternVL-4B~\cite{chen2024internvl} & 62.40 & 72.63 & 39.82 \\
    InternVL-8B~\cite{chen2024internvl}  & 63.08 & 72.93 & 39.13 \\
    Qwen2.5-VL-3B~\cite{Qwen2.5-VL} & 63.24 & 73.16 & 38.88 \\
    Qwen2.5-VL-7B~\cite{Qwen2.5-VL} & 66.72 & 75.99 & 35.86 \\
    \bottomrule
    \end{tabular}}
    \caption{More Choices of MLLM.}
\label{tab:ablation-mllm}

\end{table}

\end{document}